\documentclass[letterpaper, 10 pt, conference]{ieeeconf}  

\IEEEoverridecommandlockouts                              

\usepackage[T1]{fontenc}
\usepackage{graphicx}
\usepackage[dvipsnames]{xcolor}
\usepackage{soul}
\usepackage{booktabs}
\usepackage{gensymb}
\usepackage{amsmath}

\DeclareRobustCommand{\,}{%
   \relax\ifmmode\mskip\thinmuskip\else\thinspace\fi
}
\def\thinspace{\kern .16667em }

\title{\LARGE \bf Volitional Control of the Paretic Hand Post-Stroke Increases Finger Stiffness and Resistance to Robot-Assisted Movement}

\author{Ava Chen$^{1,\ast}$, Katelyn Lee$^{1,\ast}$, Lauren Winterbottom$^{2}$, Jingxi Xu$^{3}$, Connor Lee$^{1}$,\\ Grace Munger$^{1}$, Alexandra Deli-Ivanov$^{1}$, Dawn M. Nilsen$^{2,4}$, Joel Stein$^{2,4}$, Matei Ciocarlie$^{1,4}$
\thanks{This work was supported in part by the National Institutes of Health: National Institute of Neurological Disorders and Stroke under grant R01NS115652; Eunice Kennedy Shriver National Institute of Child Health and Human Development under award F31HD111301.}%
\thanks{\llap{\textsuperscript{$\ast$}}{These authors contributed equally to this work.}}%
\thanks{\llap{\textsuperscript{1}}{Department of Mechanical Engineering, Columbia University, New York, NY 10027, USA.}}%
\thanks{\llap{\textsuperscript{2}}{Department of Rehabilitation and Regenerative Medicine, Columbia University Irving Medical Center, New York, NY 10032, USA.}}%
\thanks{\llap{\textsuperscript{3}}{Department of Computer Science, Columbia University, New York, NY 10027, USA.}}%
\thanks{\llap{\textsuperscript{4}}{Co-Principal Investigators}}%
\thanks{\noindent Corresponding authors: \footnotesize{\texttt{(ava.chen, katelyn.lee)@columbia.edu}}}%
}

\begin{document}

\maketitle
\thispagestyle{empty}
\pagestyle{empty}

\begin{abstract}
Increased effort during use of the paretic arm and hand can provoke involuntary abnormal synergy patterns and amplify stiffness effects of muscle tone for individuals after stroke, which can add difficulty for user-controlled devices to assist hand movement during functional tasks. We study how volitional effort, exerted in an attempt to open or close the hand, affects resistance to robot-assisted movement at the finger level. We perform experiments with three chronic stroke survivors to measure changes in stiffness when the user is actively exerting effort to activate ipsilateral EMG-controlled robot-assisted hand movements, compared with when the fingers are passively stretched, as well as overall effects from sustained active engagement and use. Our results suggest that active engagement of the upper extremity increases muscle tone in the finger to a much greater degree than through passive-stretch or sustained exertion over time. Potential design implications of this work suggest that developers should anticipate higher levels of finger stiffness when relying on user-driven ipsilateral control methods for assistive or rehabilitative devices for stroke.
\end{abstract}
\vspace{-4mm}

\section{INTRODUCTION}
Loss of manual dexterity due to diminished motor control and weakness in the hand, in particular limited capacity to extend the fingers for releasing a grasp, is a prevalent and debilitating impairment after stroke~\cite{barry2022, kamper2006}. Recent advances in wearable robotics have demonstrated ability to assist hand strength and range of motion to complete functional tasks~\cite{khalid2023}. By motivating and enabling the user to use the impaired limb in the process of performing daily activities, wearable robotic devices have the potential to counteract learned non-use and prevent further deterioration of motor function~\cite{park2020, ballester2016}.

However, increased effort during arm use can provoke involuntary abnormal synergy patterns, leading to increases in muscle tone and in motor dysfunction~\cite{mcpherson2022, pundik2019}.
Such complications increase finger stiffness, which directly interferes with the ability for robotic devices to assist hand movement. Thus, monitoring and adapting to changes in muscle tone during volitional hand movement is critical for developing safe and effective \mbox{robotic interventions.}

Quantifying changes in muscle tone, often by modeling their effects on movement with stiffness parameters~\cite{cha2020}, has long been a focus of research for stroke rehabilitation.
Multiple groups have monitored muscle tone while the fingers are passively stretched~\cite{kanosue1983, kamper2000, wang2021}, including some that investigated the dependency of stiffness fluctuation with speed of robotic pertubation~\cite{zhang2019, cruz2010}. Ranzani et al. embedded resistive stiffness measurements into a robotic therapy regimen to monitor the evolution of muscle tone over the course of a session~\cite{ranzani2023}. Others have studied effects from attempting to move the limb even when motor dysfunction prevents execution of movement~\cite{wang2019}. McPherson and Dewald used isometric exercises to observe increased muscle tone distally during maximal muscular contractions at the proximal upper limb~\cite{mcpherson2022, mcpherson2019}. However, prior methods have only studied muscle tone when the hand is passively moved or does not move. A significant gap remains in our understanding of the effects of muscle tone in the context of \textit{active hand-arm movement under volitional control}, especially for users who require assistance to fully open the hand.

\begin{figure}[t]
    \centering
    \vspace{2.5mm}
    \includegraphics[width=.96\columnwidth]{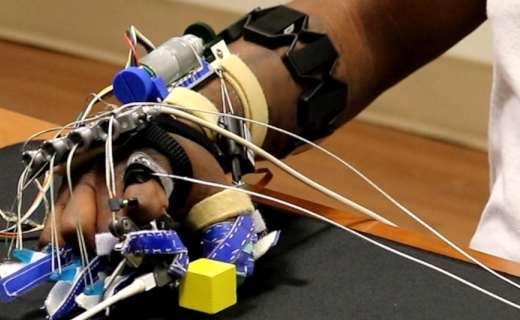}
    \vspace{-3mm}
    \caption{Subject engaging the impaired limb to practice functional hand movements with robotic assistance. Volitional ipsilateral activation of the orthosis makes user-driven, intensive exercise possible for individuals who cannot open their own hands.}
    \label{fig:myhand}
    \vspace{-6mm}
\end{figure}

In this study, we examine how volitional effort, exerted in an attempt to open or close the hemiparetic hand, affects resistance to robot-assisted movement, modeled as stiffness. To achieve this, we introduce a wearable orthosis capable of simultaneously actuating fingers for functional tasks and measuring applied force and resultant finger movement. Working with three chronic stroke subjects, we measure stiffness fluctuations of
the index finger when the user is actively exerting effort to move the hand with robotic assistance, compared with when the user is relaxed and allows the hand to be passively stretched. We also monitor the evolution of stiffness over the course of the session, to compare against general effects of sustained active engagement and arm use over time. Main contributions of this work include:
\begin{itemize}
    \item To the best of our knowledge, we introduce the first continuous measurements of finger resistance to movement in response to externally-provided assistive force \textit{that is activated by voluntary muscle recruitment at the impaired limb}. These observations allow us to study effects of active engagement (attempt at volitional control) on resistance to movement for an assisted \mbox{paretic hand.}
    \item We observe that active motor recruitment of the upper limb during functional use increases muscle tone in the finger to a much greater degree than can be explained by sustained exertion over time or velocity-dependent passive stretch. 
    \item Our results have potential design implications, as they suggest that developers should anticipate higher magnitudes of joint stiffness when relying on user-driven ipsilateral control methods for assistive or rehabilitative devices for stroke.
\end{itemize}
\section{METHODS}

This study uses the MyHand, a wearable robotic hand orthosis for hemiparetic assistance and rehabilitation, as a sensing and movement-completion platform~(Figure~\ref{fig:myhand}). We developed this device with the goal of harnessing hemiparetic users' residual hand function to provide hand-opening assistance while allowing the hand to close under body power. The device uses an underactuated tendon-network winch mechanism to extend all fingers simultaneously with a single motor. These tendon networks are anchored to the fingertips with 3D-printed components that splint the distal interphalangeal joints; motorized retraction of the tendons transmits finger-extension torques to the MCP and PIP of each finger, opening the hand. The anchoring structures of the fingersplint components and motor force profile were designed~\cite{park2018} to overcome the expected finger stiffnesses of a user with moderate spasticity, as reported by Kamper et al~\cite{kamper2000}. A second tendon-network extends and stabilizes the thumb for oppositional grasping~\cite{chen2022}. Motors are mounted to an aluminum splint that fixes the wrist at a neutral angle, and the device is secured to the body with velcro straps around the digits, hand, and arm. Dimensions of the fingertip components, lengths of each tendon, and motor setpoints for opening/closing the hand are fully customized for each user. The total weight of the device with sensor cables (excluding tabletop electronics and power supply) is 371 grams.

\begin{figure}[t]
    \centering
    \vspace{.25cm}
    \includegraphics[trim={0 0 4mm 2mm},clip,width=\columnwidth]{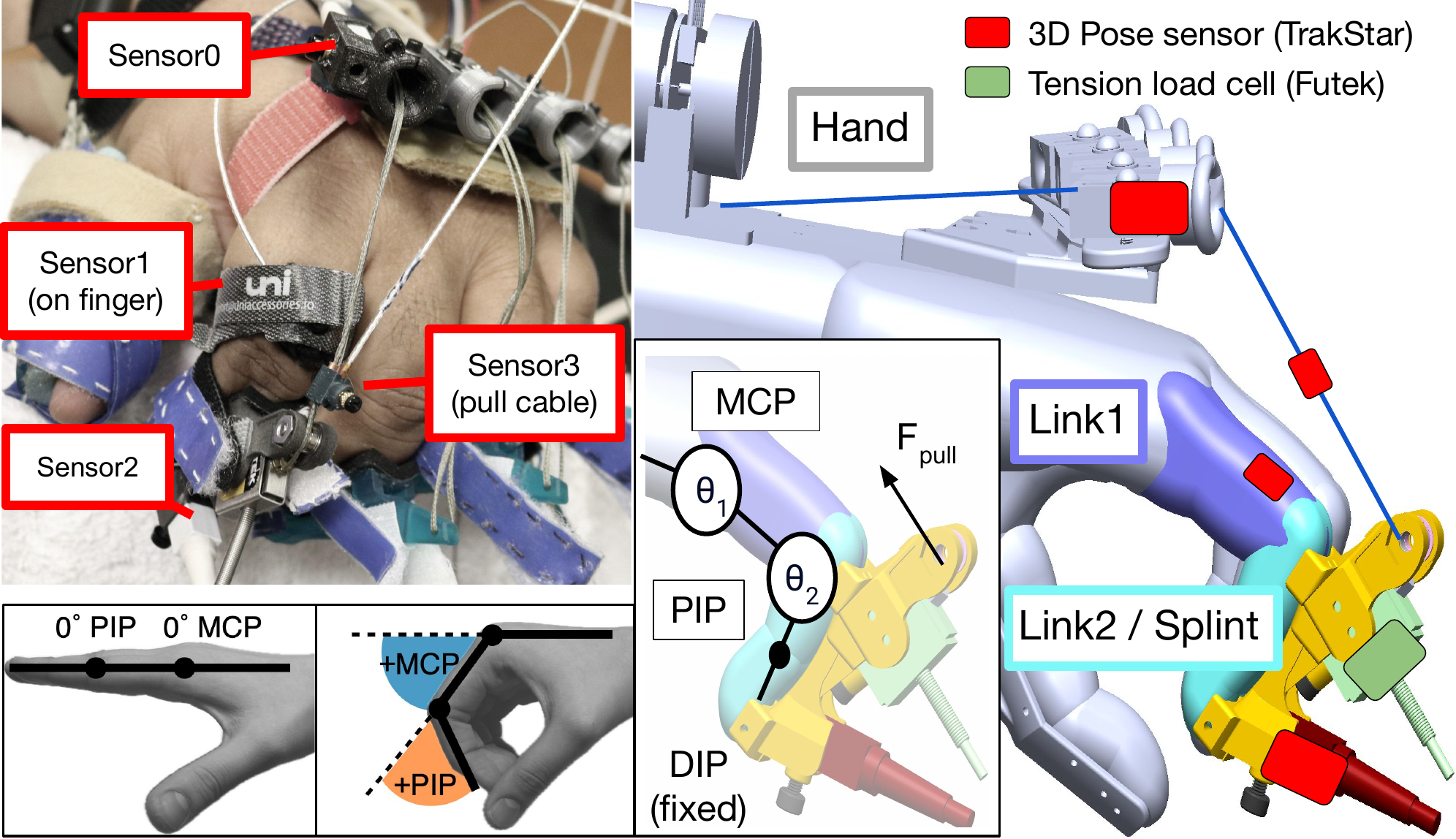} 
    \vspace{-7mm}
    \caption{Diagrams showing sensor placements on the robot and hand and the angle convention used in this paper.}
    \label{fig:sensorplacement}
    \vspace{-6mm}
\end{figure}

\begin{figure*}[t]
    \centering
    \vspace{.25cm}
    \includegraphics[width=0.8\textwidth]{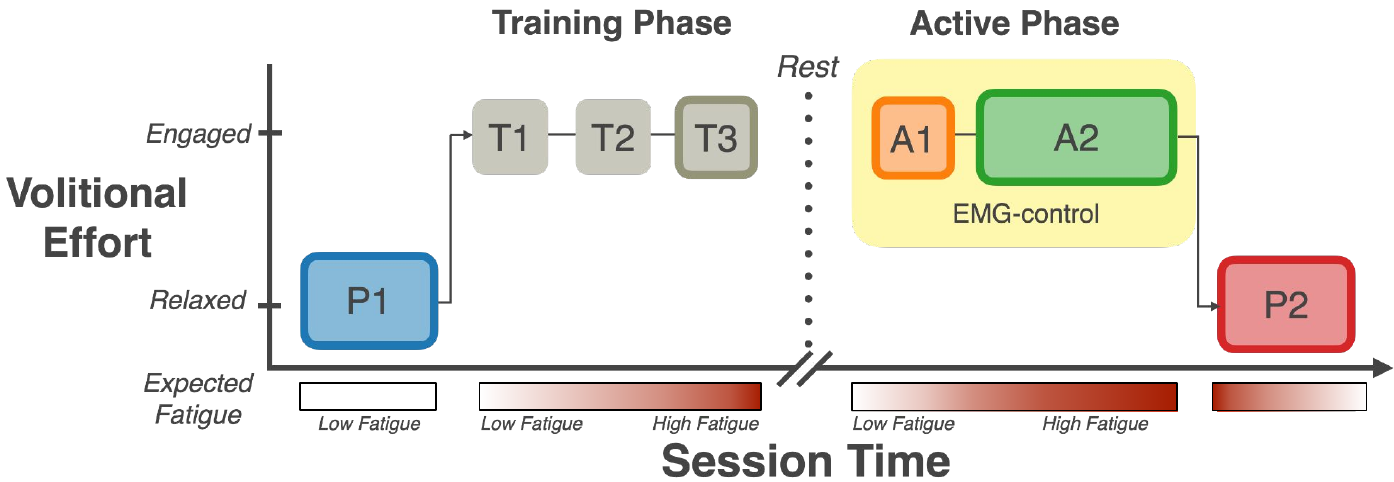}
    \vspace{-2mm}
    \caption{Graphical summary of the experimental protocol. The session is divided into four main phases: an initial passive phase (P1), a training phase (T1-T3), an active phase (A1-A2) , and a final passive phase (P2). During the passive phases (P1, P2), we use button control to open and close the participant's hand. This contrasts with the active phase (A1, A2), in which the participant has full volitional control of the orthosis via EMG. Our EMG-control is calibrated during the training phase in which the participant is actively exerting volitional effort to move their hand both without (T1-T2) and with (T3) assistance from the orthosis. Bolded outlines around an experimental phase block denote when the orthosis assists hand movement. The gradient color bar labeled ``Expected Fatigue'' at the bottom of the diagram qualitatively denotes our expectations of increasing fatigue (red) over the course of the session as the participant engages in activities of sustained active engagement and use of the arm.}
    \label{fig:protocol}
    \vspace{-5mm}
\end{figure*}

MyHand employs two methods for activating robotic hand opening/closing. The first method acts on volitional effort from the user, for which the robot detects user intent to open, close, or relax the hand from ipsilateral forearm surface electromyography (EMG) signals while the user attempts to perform hand movements. Our approach for EMG intent inferral is described in detail in previous work with chronic stroke survivors~\cite{park2020, xu2022}, and uses signals from a commercial EMG armband (Myo, Thalmic Labs) that consists of eight sensors distributed around the arm circumference. The second method allows for the user to remain relaxed and exert no effort while the hand is passively opened and closed. In this second method, researchers send robot commands to open or close the hand by pushing a button.

\subsection{Muscle tone estimation}
In this study, we measure effects of muscle tone by evaluating the change in resistance to externally-applied force as the fingers are extended. We measure stiffness by evaluating the change in cable tension per cable retraction distance, which gives us an estimate of how much resistance is overcome for the robot to assist hand-opening.

We obtain applied force readings for the index finger by mounting a uniaxial load cell (FSH00097, Futek, force \mbox{resolution 0.28N)} within the fingertip splint in series with the finger-specific pull-tendon. The sensorized splint incorporates a bearing-mounted pulley to minimize sensing losses due to possible off-axis moments during contact with objects. Cable retraction distance is measured by the motor encoder since all fingers move together.

We obtain stiffness measurements by fitting a slope to the cable force-displacement data via linear regression on the points between the first instance of force above the 0.28N sensor threshold and the maximum peak force, for each hand-opening movement, within a given experimental phase. 

\subsection{Finger Joint Tracking}
We track relative MCP and PIP joint rotations during active and passive finger extension movements. Four electromagnetic sensors (Models 800 and 180, Northern Digital) placed along the index finger's pull-tendon transmission collect 3D orientation data. We integrate these sensors into the orthosis at the fingertip splint, at the cable guide above the MCP, clamped directly onto the pull-tendon cable, and strapped to the proximal phalanx. (See Figure~\ref{fig:sensorplacement} for a diagram of sensor locations.) The manufacturer's sensor-fusion unit (3D Guidance \mbox{trakSTAR}, Northern Digital) tracks each sensor's position and orientation from a stationary transmitter and records data as a time-synchronized transform. To minimize error from possible device migration during the session, between each experimental phase of the session we manually sweep the index finger through its full passive range of motion. We obtain the axis of rotation for each joint by taking the normal direction to a plane fitted to the fingertip positions recorded during the sweeps. We use the swing-twist decomposition method~\cite{dobrowolski2015} to obtain the component of each sensor rotation around the joint axis in order to extract MCP and PIP joint angles. We empirically determined average RMS angle error with our system to be $1.2^{\circ}$. 

\subsection{Pilot Clinical Evaluation}

This work studies a common subset of the chronic hemiparetic stroke population who retain the ability to actively flex their fingers to perform a gross grasp, but are unable to volitionally extend their fingers to release the grasp. All research participants provided informed consent to participate in the study, which was conducted in accordance with the protocol (IRB-AAAS8104) approved by the Columbia University Medical Center Institutional Review Board. Subjects had prior experience with earlier prototypes of the robotic device and control methods, and were recruited as part of a device investigation study (NCT04436042) for stroke. Experiments were performed in a clinical setting under supervision of an occupational therapist.

We performed experiments with three chronic stroke survivors having hemiparesis and limited upper-limb motor function as measured by Fugl-Meyer score (FM-UE). Subjects met the following inclusion criteria: (1) at least 18 years of age; (2) at least 6 months post-stroke; (3) muscle tone and spasticity scoring MAS $\leq 2$ in digits, wrist, and elbow; (4) passive range of motion of digits and wrist within functional limits; (5) inability to fully extend fingers without assistance; (6) sufficient active flexion in arm and hand to form a closed fist and lift the arm above table height; (7) intact cognition to provide informed consent and follow complex instructions. Participant characteristics are summarized in Table~{\ref{tab:baseline_table}}.

\begin{table}[h]
	\caption{Participant Characteristics}
	\centering
	\label{tab:baseline_table}
	\begin{tabular}{l| c c c}
		Subject & S1 & S2 & S3\\
		\hline \hline
        Gender & M & F & M\\
		Age & 48 & 43 & 50\\
		Years Since Stroke & 10 & 8 & 10\\
        Affected Side & R & L & L\\ \hline
		FM-UE Total & 27 & 29 & 26\\ 
        \hspace{3mm} Hand Subscore & 2 & 2 & 1 \\\hline
		MAS &  &  & \\	
            \hspace{3mm} Elbow Flexor & 2 &\ 1+ & 2\\
		\hspace{3mm} Index Finger Flexor & \ \ 1+ & 1 &\ \ 1+
	\end{tabular}	
 \vspace{-6mm}
\end{table}

\subsection{Experimental Protocol}
The experimental protocol consists of a 90-min robotic training session that includes phases of \textit{passive hand movement}, in which the subject relaxes the hand and passively allows the robot to extend the fingers, and \textit{active hand movement}, in which the subject actively attempts to open the hand by exerting voluntary muscle contractions to control the robot with EMG signals. Velocity of robot-assisted hand opening/closing remains constant across all subjects and experimental conditions. The subject is seated at a table with the forearm supported on a towel roll for all experimental phases except for the functional cube task. Setup time at the start of the session includes approximately 15 minutes to assess spasticity with the Modified Ashworth Scale (MAS), put on the device, and adjust hardware for comfort and fit. The active hand movement phases includes exercises that we have adapted from previous studies with subjects with stroke using EMG-control and the MyHand device~\mbox{\cite{park2020, xu2022}.} Figure~\ref{fig:protocol} shows a summary schematic of the protocol.

\begin{figure*}[t]
    \centering
    \vspace{.25cm}
    \includegraphics[width=.99\textwidth]{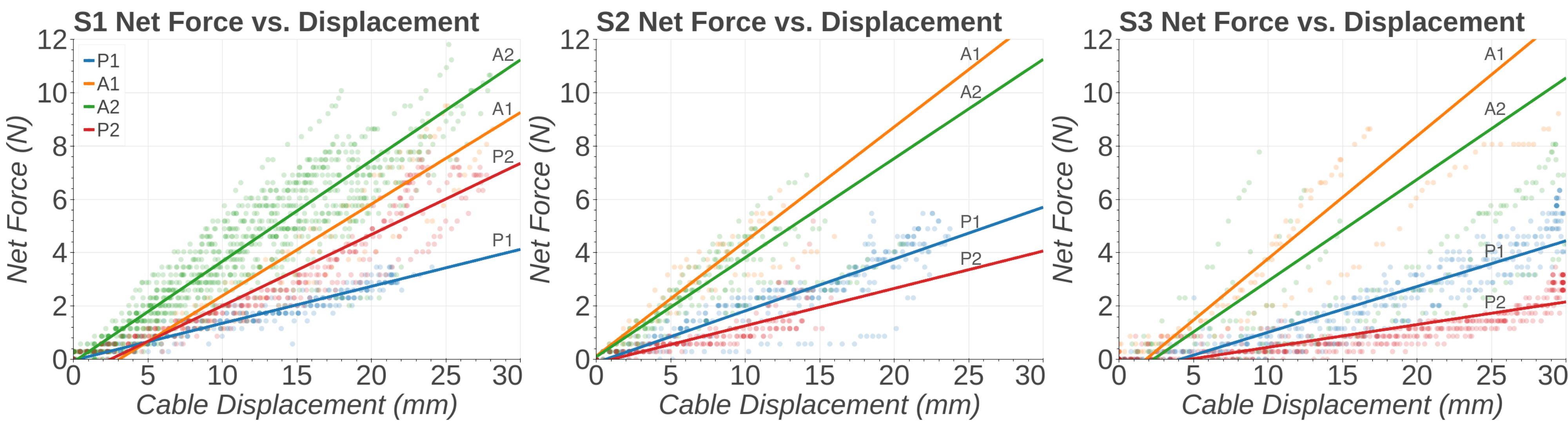}
    \vspace{-3mm}
    \caption{Net force versus motor displacement scatterplots and corresponding best fit linear regression lines for each experimental condition during robot-assisted hand openings across subjects. Experimental conditions are labeled by color (Blue = P1, Orange = A1, Green = A2, Red = P2).}
    \label{fig:force}
    \vspace{-2mm}
\end{figure*}

\begin{table*}[t]
    \renewcommand{\arraystretch}{1.3}
    \caption{Average finger stiffness values and standard deviation for 3 stroke subjects. We also report the average stiffness across all experimental conditions. Bold-text highlights the highest stiffness result for each experimental condition.}
    \vspace{-2mm}
    \label{tab:stiffness}
    \centering
    \begin{tabular}{c|cccc|cc}
    \toprule
    \textbf{Participant} & 
    \textbf{P1 Phase} &  \textbf{A1 Phase} & \textbf{A2 Phase} & \textbf{P2 Phase}&
    \begin{tabular}[c]{@{}c@{}}\textbf{Passive}\\\textbf{Average} \end{tabular} & 
    \begin{tabular}[c]{@{}c@{}}\textbf{Active}\\\textbf{Average} \end{tabular} \\
    \midrule
    S1 & 0.139 $\pm$ 0.009 & 0.344 $\pm$ 0.060 & \textbf{0.378 $\pm$ 0.092} & 0.267 $\pm$ 0.030 & 0.203 $\pm$ 0.022 & \textbf{0.361 $\pm$ 0.091}\\
    
    S2 & 0.194 $\pm$  0.029 & \textbf{0.430 $\pm$ 0.103} & 0.372 $\pm$ 0.130 & 0.140 $\pm$ 0.063 & 0.167 $\pm$ 0.049 & \textbf{0.401 $\pm$ 0.125}\\
    
    S3 & 0.172 $\pm$ 0.025 & \textbf{0.460 $\pm$ 0.131} & 0.381 $\pm$ 0.249 & 0.085 $\pm$ 0.017 & 0.128 $\pm$ 0.021 & \textbf{0.421 $\pm$ 0.222}\\
    \midrule
    
    Average & 0.168 $\pm$ 0.022 & \textbf{0.411 $\pm$ 0.102} & 0.377 $\pm$ 0.128 & 0.164 $\pm$ 0.041 & 0.166 $\pm$ 0.033 & \textbf{0.394 $\pm$ 0.125}\\
    \bottomrule
    \end{tabular}
    \vspace{-2mm}
\end{table*}

To characterize the change in finger stiffness between passive and active movement, we divide the session into several phases of varying activity. The first phase is the baseline passive movement phase, P1, in which we control the orthosis via button-press to open and close the subject's hand. We perform ten total open/close movements over an approximate 2-min timespan, with a 1-sec pause between each open and close. Because the subject's hand is passively opened by the orthosis, we can determine their baseline joint resistance to passive movement. Following this phase, we begin the training phase, T1-T3, in which the the subject is guided through a sequence of verbal commands to \textit{relax, open, relax, close, and finally relax} their hand to enable EMG-control of the orthosis. Within each phase, this sequence is repeated for three total rounds. The training T1-T3 follows the classifier training protocol that we typically use to calibrate EMG control: there are two blocks (T1-T2) during which the robot does not assist movement, followed by one block (T3) in which we use button-control to move the motor approximately one second after the verbal command is given. The subject is allowed to take an optional 1-2 min. break between blocks. The total training phase T1-T3 takes approximately \mbox{15-20 min.} A 5-min rest period follows the final block T3 in order to minimize potential effects of fatigue introduced during training.

Then, we begin the active phases, A1 and A2, in which the orthosis is under full user EMG-control. This means the orthosis will open when the classifier detects volitional intent to open their hand. The first phase, A1, follows the training protocol sequence of opening, closing, and relaxing the hand to provide a direct comparison of a verbally-cued open command. For the second phase, A2, we ask the subject to pick up a target cube and place it in our hand. The main intent of the functional cube task is to motivate the subject to sustain active engagement of the arm and impaired hand to collect many open/close cycles. It also serves as an activity intended to induce fatigue and increase demand on the proximal arm. The A2 phase continues \textit{for as long as session time allows}, approximately 15-25 min. The final phase of the experiment is another passive movement phase, P2, in which we repeat the P1 protocol consisting of ten passive opens under button-control. Collecting passive data at the beginning and end of the session allows us to observe effects from the active phase.

\section{RESULTS AND DISCUSSION}

We analyze index finger extension movements from phases A1 and A2 to compare hand movements under volitional control, or \textit{active movement}, against passive index finger extensions from phases P1 and P2, or \textit{passive movement.}

\begin{figure*}[t]
\centering
\vspace{1cm}
\begin{minipage}[t]{0.48\textwidth}
    \centering
    \includegraphics[height=0.72\textheight]{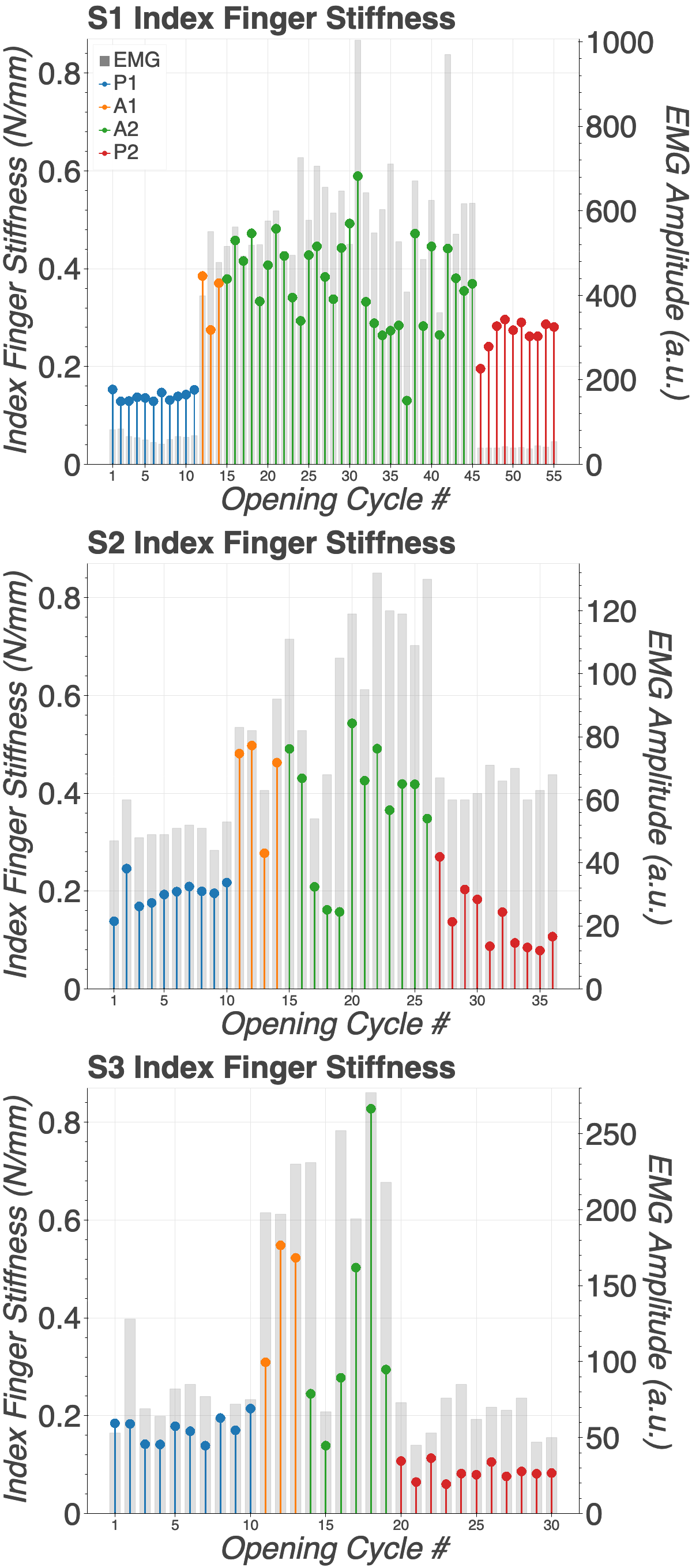}
    \vspace{-3mm}
     \caption{Spike plot of measured index finger stiffnesses ordered chronologically during robot-assisted hand openings. Experimental conditions are labeled by color (Blue = P1, Orange = A1, Green = A2, Red = P2). We define finger stiffness as the slopes of the best-fit lines calculated by performing linear regression on the force and motor displacement data during hand opening. EMG bars (Gray) depict the maximum EMG amplitude recorded during each hand opening. EMG amplitude values are integers given in arbitrary units (a.u.) transmitted by the commercial EMG armband. We set scale bars for EMG amplitude based on the maximum values recorded for each subject.}
     \label{fig:stiffness}
 \end{minipage} \hfill
 \begin{minipage}[t]{0.48\textwidth}
    \centering
    \includegraphics[height=0.72\textheight]{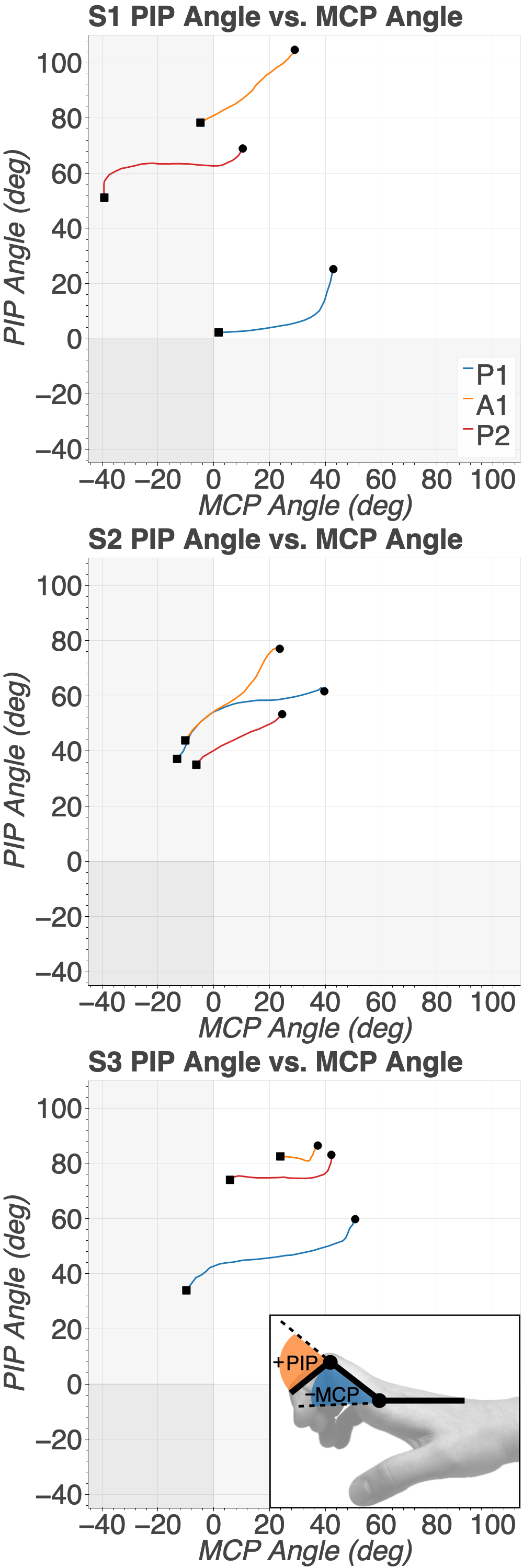}
    \vspace{-3mm}
     \caption{Representative joint angle trajectories of index finger MCP and PIP during robot-assisted hand openings in passive phases P1, P2, and first active phase A1. Pictured is the third hand opening for each experimental condition (Blue = P1, Red = P2, Orange = A1). A black circle marks the initial resting angle (hand closed), and a black square marks the final angle (hand opened). A zero angle value corresponds to full joint extension. 
     Gray regions mark negative angle values, corresponding to PIP and MCP hyperextension. The image on the bottom right depicts the ``claw'' posture we observe in Subjects S1 and S3, where the finger has a positive PIP joint angle and negative MCP joint angle, corresponding to MCP hyperextension.}
     \label{fig:angledisp}
 \end{minipage}
 \vspace{-0.4cm}
\end{figure*}

\subsection{Finger stiffness response with active hand use}
To calculate stiffness, we perform linear regression on the net force and motor retraction displacement at the index finger during each hand opening. Figure~\ref{fig:force} plots average best-fit lines and each recorded force-displacement measurement for each instance of hand opening for all experimental conditions P1, A1, A2, and P2. Figure~\ref{fig:stiffness} shows the computed stiffnesses taken from each hand-opening cycle with the maximum EMG amplitude reached during each hand-opening. Table~\ref{tab:stiffness} presents average finger stiffnesses and errors across subjects and experimental conditions. The active phases had the highest recorded finger stiffnesses for all subjects, for which the highest recorded average finger stiffness value is 0.46 N/mm from Subject S3 during phase A1. 

Figure~\ref{fig:stiffness} shows the calculated finger stiffness value plotted with the maximum EMG magnitude recorded during each hand opening. Pairing the finger stiffness value with the maximum EMG magnitude allows us to observe a general trend of greater maximum EMG magnitude during the active phases than the passive phases. This aligns with the corresponding increase in muscle activity that would accompany active movements, rather than passive movements. 

\subsection{Evolution of joint angle trajectories over session time}
Having measured MCP and PIP joint angles throughout the session, we observe changes in index finger joint resting pose and joint angle trajectory. Figure~\ref{fig:angledisp} compares the joint angle trajectories between the two passive conditions that flank the session (P1, P2) and the first active phase (A1). Subjects S1 and S3 demonstrate a dramatic change in resting angle pose for the PIP joint between the passive phases, P1 and P2. For Subject S1, we observe that the PIP and MCP joints' initial full extension to 0$\degree$ at the beginning of the session (P1) shifts to reduced PIP joint motion and MCP hyperextension to $-40\degree$ by the end of the session (P2). In concert with this, we note that the resting angle of the PIP joint for Subject S1 between the P1 and P2 phases shifts from $20\degree$ to $70\degree$, demonstrating a hand pose during robot-assisted hand opening we call ``claw hand'' (bottom-right of Figure~\ref{fig:angledisp}). Both Subjects S1 and S3 first adopt the ``claw hand'' posture during A1, the first instance of active movement, which suggests that active, volitional control contributes to this change in finger and hand posture that hinders full finger extension and effective robot-assisted hand opening. 

\subsection{Evolution of joint stiffness over session time}
Despite variations across the three subjects in overall muscle tone levels, the magnitude differences between phases of the experiment, and effects of fatigue, we consistently observe higher joint stiffnesses during active tasks than passive tasks. Our effects are from volitional exertion, not just fatigue over time. Comparing within the two active phases (A1, A2), we see that there is no clear rise in stiffness within the three iterations of \textit{open, relax, close} of A1, or a general rising trend within the prolonged intensive activity phase of A2. This is more evidence suggesting that fatigue is not the main contributor to the increase in finger resistance.

\subsection{Future work and Limitations}
This work offers interesting directions toward developing novel devices that account for the increase in joint-stiffness during active, functional tasks. Robotic devices that already utilize a mix of rigid and soft components to apply external force to the body are ideal platforms for mounting force- and orientation-tracking sensors, which then allow the assistive device to perform a dual role of monitoring therapeutic outcomes in contexts of functional movement and hand use. 

Several factors present limitations to this pilot study. The limited subject sample size and subset of tasks prevents us from drawing conclusions about possible mechanisms (spasticity, abnormal synergies, fatigue, etc.) that might contribute to our observed stiffness effects. Future investigation is needed to determine whether our observed stiffness effects are indeed intrinsic biomechanical observations of muscle tone, or if they are more attributable to device efficiency losses that are magnified by PIP resistances in particular.
\section{CONCLUSIONS}
We present observations of index finger resistance to movement, modeled as stiffnesses, in the context of \textit{active hand and arm movement under volitional control}. Comparisons of finger-level stiffnesses between the active and passive movement phases of the experimental protocol suggests that active motor recruitment of the upper limb in an attempt to move the hand is the main contributor to increased muscle tone in the finger during robot-assisted hand movement. To our knowledge, these are the first observations of finger-level stiffness effects during hand movements controlled by active ipsilateral engagement of the paretic limb. Our observations highlight multiple possible directions for future research and offer potential design implications for device developers, including updated expectations of assisted-movement effects of muscle tone during active use of the hand.


\end{document}